%% file: main.tex
\documentclass{article}

\usepackage{microtype}
\usepackage{graphicx}
\usepackage{subfigure}
\usepackage{booktabs}

\usepackage{hyperref}
\usepackage{makecell}

\usepackage[accepted]{icml2025}

\usepackage{amsmath}
\usepackage{amssymb}
\usepackage{mathtools}
\usepackage{amsthm}

\usepackage[capitalize,noabbrev]{cleveref}
\usepackage{diagbox}
\usepackage{subcaption}
\usepackage{graphicx}
\usepackage{listings}
\usepackage{float}
\usepackage{placeins}
\usepackage{listings}
\usepackage{xcolor}

\theoremstyle{plain}

\theoremstyle{definition}

\theoremstyle{remark}

\newif\ifdraft
\draftfalse

\input{macros}
\input{math_commands}

\icmltitlerunning{Implementing Adaptations for Vision AutoRegressive Model}
\begin{document}

\twocolumn[
\icmltitle{Implementing Adaptations for Vision AutoRegressive Model}

\icmlsetsymbol{equal}{*}

\begin{icmlauthorlist}
\icmlauthor{Kaif Shaikh}{cispa}
\icmlauthor{Franziska Boenisch}{cispa}
\icmlauthor{Adam Dziedzic}{cispa}
\end{icmlauthorlist}

\icmlaffiliation{cispa}{CISPA Helmholtz Center for Information Security, Saarbr\"ucken, Germany}

\icmlcorrespondingauthor{Kaif Shaikh}{kaif.shaikh@cispa.de}
\icmlcorrespondingauthor{Franziska Boenisch}{boenisch@cispa.de}
\icmlcorrespondingauthor{Adam Dziedzic}{dziedzic@cispa.de}

\icmlkeywords{Vision ARs, Machine Learning, Differential Privacy}

\vskip 0.3in
]

\printAffiliationsAndNotice{} 

\input{sec/00abstract}
\input{sec/01intro}
\input{sec/02background}
\input{sec/03nondp_finetuning}
\input{sec/04dp_finetuning}
\input{sec/05conclusions}

\input{sec/10ack}

\bibliographystyle{plainnat}
\bibliography{main}

\clearpage
\onecolumn
\appendix
\input{sec/20appendix}

\end{document}

%% file: macros.tex
\usepackage{xparse}
\usepackage{xspace}
\usepackage{algorithm}%
\usepackage{fixltx2e}
\usepackage{tikz}
\usepackage{float}
\usepackage{multirow}
\usepackage{multicol}
\usepackage{colortbl}
\usepackage{array}
\newcommand{\PreserveBackslash}[1]{\let\temp=\\#1\let\\=\temp}
\newcolumntype{C}[1]{>{\PreserveBackslash\centering}p{#1}}
\newcolumntype{R}[1]{>{\PreserveBackslash\raggedleft}p{#1}}
\newcolumntype{L}[1]{>{\PreserveBackslash\raggedright}p{#1}}

\usepackage{microtype}
\usepackage{graphicx}
\usepackage{mathtools}
\usepackage{float}
\usepackage{subcaption}
\usepackage{booktabs} %
\usepackage[shortlabels]{enumitem}

\usepackage{amssymb}%
\usepackage{pifont}%

\usepackage{bbold}

\usepackage{hyperref,amssymb,enumitem}

\setlist[itemize]{leftmargin=*}
\setlist[enumerate]{leftmargin=*}

\newcommand{\cmark}{\ding{51}}%
\newcommand{\xmark}{\ding{53}}%

\newcommand*{\rej}{{\ooalign{\lower.3ex\hbox{$\sqcup$}\cr\raise.4ex\hbox{$\sqcap$}}}}

\usepackage{stackengine}

\usepackage{xcolor}

\newcommand{\ie}{\textit{i.e.,}\@\xspace}
\newcommand{\eg}{\textit{e.g.,}\@\xspace}

\graphicspath{{images/}}

\usepackage{booktabs,arydshln}
\makeatletter
\def\adl@drawiv#1#2#3{%
        \hskip.5\tabcolsep
        \xleaders#3{#2.5\@tempdimb #1{1}#2.5\@tempdimb}%
                #2\z@ plus1fil minus1fil\relax
        \hskip.5\tabcolsep}
\newcommand{\cdashlinelr}[1]{%
  \noalign{\vskip\aboverulesep
           \global\let\@dashdrawstore\adl@draw
           \global\let\adl@draw\adl@drawiv}
  \cdashline{#1}
  \noalign{\global\let\adl@draw\@dashdrawstore
           \vskip\belowrulesep}}
\makeatother

\usepackage{cleveref}
\crefalias{prop}{proposition} %

\newcommand{\nlp}[1]{}

\newcolumntype{x}[1]{>{\centering\arraybackslash\hspace{0pt}}p{#1}}

\usepackage{amsmath}
\usepackage{xspace}

\ifdraft
\usepackage[textsize=tiny]{todonotes}

\newcommand{\mytodo}[1]{\textcolor{red}{[todo: #1]}}
\newcommand{\mycomment}[1]{\textcolor{red}{[comment: #1]}}
\newcommand{\antoni}[1]{\textcolor{magenta}{AK: #1}}
\newcommand{\adam}[1]{\textcolor{cyan}{[AD: #1]}}
\newcommand{\franzi}[1]{\textcolor{brown}{FB: #1}}

\else
\usepackage[disable]{todonotes}
\newcommand{\franzi}[1]{}
\newcommand{\adam}[1]{}
\newcommand{\matthew}[1]{}
\newcommand{\dariush}[1]{}
\newcommand{\mytodo}[1]{}
\newcommand{\mycomment}[1]{}
\newcommand{\antoni}[1]{}
\fi

%% file: math_commands.tex
\usepackage{amsmath,amsfonts,bm}

\def\eqref#1{equation~\ref{#1}}

\def\1{\bm{1}}

\DeclareMathAlphabet{\mathsfit}{\encodingdefault}{\sfdefault}{m}{sl}
\SetMathAlphabet{\mathsfit}{bold}{\encodingdefault}{\sfdefault}{bx}{n}



%% file: sec/00abstract.tex
\begin{abstract}
Vision AutoRegressive model (VAR) was recently introduced as an alternative to Diffusion Models (DMs) in image generation domain.
In this work we focus on its adaptations, which aim to fine-tune pre-trained models to perform specific downstream tasks, like medical data generation. While for DMs there exist many techniques, adaptations for VAR remain underexplored. Similarly, differentially private (DP) adaptations---ones that aim to preserve privacy of the adaptation data---have been extensively studied for DMs, while VAR lacks such solutions. 
In our work, we implement and benchmark many strategies for VAR, and compare them to state-of-the-art DM adaptation strategies. We observe that VAR outperforms DMs for non-DP adaptations, however, the performance of DP suffers, which necessitates further research in private adaptations for VAR. Code is available at \url{https://github.com/sprintml/finetuning_var_dp}.
\end{abstract}

%% file: sec/01intro.tex
\section{Introduction}
Recently, Vision AutoRegressive model (VAR)~\citep{VAR} has been proposed as a powerful alternative to Diffusion Models (DMs)~\citep{rombach2022high} in image generation. 
Yet, while for DMs, there exist multiple strong methods~\citep{xie2023difffitunlockingtransferabilitylarge,ruiz2023dreambooth,textualinversion} to adapt pre-trained models to specific downstream tasks, like medical data generation~\citep{dmsinmedical}, similar adaptations for VAR remain underexplored. 
Our work makes the first step in understanding and evaluating possible adaptations, ranging from full- to parameter-efficient fine-tuning (LoRA~\citep{hulora}, LayerNorm~\citep{zhao2023tuninglayernormattentionefficient}), by implementing and benchmarking the methods on the class-conditioned VAR. 

Since the adaptation data might consist of highly-sensitive samples, it is crucial that fine-tuned models do not leak privacy. One of the methods to prevent the leakage employs Differential Privacy (DP)~\citep{dwork2006} to protect the vulnerable data. We explain how to overcome limitations of the VAR code base to implement privacy-preserving adaptations, then we implement and benchmark them.

We compare the performance of our adaptations of VAR to the existing SOTA DM adaptation---DiffFit---on five downstream datasets. We find that VAR adapts faster, is computationally more efficient, and outperforms DiffFit in generation quality. Yet, DP-adaptations of VAR suffer from low generation quality, as well as slow convergence. We revisit augmentation multiplicity~\citep{de2022unlocking}, a strategy that improves the performance of DP models at a cost of higher compute, and find that it is beneficial for DP fine-tuning. However, the performance of DP-adaptations remains low, which prompts for further work in that domain.

Our released code with implementation of the methods and benchmarking serves to aid the researchers and practitioners to build and evaluate novel adaptation methods, and to close the gap between DMs and VARs in that domain.
\label{sec:intro}

%% file: sec/02background.tex
\section{Background and Related Work}
\label{sec:background}
\textbf{Image AutoRegressive models (IARs)} are a class of generative models that create images by modeling 
$p(x)=\prod_{n=1}^{N}p(x_n|x_{<n})$, 
where $x=(x_1,x_2,\dots,x_N)$ is an image represented as a sequence of $N$ tokens, $x_n$ is the n-th token. First proposed by~\citet{chen2020generative} and further developed by~\citet{VAR,yu2024randomized,Infinity} they offer better generation quality than their predecessor, namely DMs~\citep{rombach2022high}. 

IARs are trained to minimize $\mathcal{L}_{\text{AR}}=\mathbb{E}_{x\sim\mathcal{D}}\left[-\text{log}\left(p(x)\right)\right]$, where $\mathcal{D}$ is a dataset of tokenized images. During generation, given a prefix $x_{<n}$ the model outputs per-token logits $p(x_n|x_{<n})$, from which we iteratively sample the next token to obtain the final image $\hat{x}$. In practice, the images are tokenized using a pre-trained VQ-GAN~\citep{esser2020taming}, and Transformer-based~\citep{attentionisallyouneed} architectures like GPT-2~\citep{radford2019language} serve as the autoregressive backbone for modeling $p(x)$.

\textbf{Vision AutoRegressive (VAR) model} is an IAR proposed by~\citet{VAR}, which inspired SOTA image generative model---Infinity~\citep{Infinity}. VAR shifts the paradigm from next-token prediction to next-scale prediction. Instead of predicting 1D token sequences in raster-scan order (top to bottom, left to right), images are processed as sequences of 2D token grids, starting from lower to higher resolution. Effectively, VAR generates images quicker than classical IARs, requiring less predictions, as the 2D token grid for each scale can be sampled simultaneously.

\textbf{Full Fine-Tuning (FFT)} is an adaptation technique that updates every parameter of a model, given a fine-tuning dataset. Contrary to regular training, it is initialized from a pre-trained model (instead of from random initialization), and the end goal of FFT is to tailor the model to a specific task, based on a small, domain-specific dataset.

\textbf{Parameter-Efficient Fine-Tuning (PEFT)} helps fine-tune a model in a more compute-efficient manner by reducing the number of trainable parameters and memory usage as compared to FFT~\citep{xu2023parameterefficientfinetuningmethodspretrained}. It is widely used in adapting general foundation models, pre-trained on large datasets, to specific downstream tasks, where FFT is not required to achieve satisfactory performance.

\textit{Low-Rank Adaptation} (LoRA)~\cite{hulora} is a PEFT method widely used for Transformer-based architectures. Such models have several dense layers with weight matrices ($\bm{W})$ of full rank. However, during fine-tuning, the updates ($\Delta\bm{W}\in\mathbb{R}^{d\times k}$) to these layers exhibit a low intrinsic rank~\citep{aghajanyan2020intrinsicdimensionalityexplainseffectiveness}. LoRA builds on that phenomenon, and attempts to decompose the update into multiplication of two low-rank matrices, $\Delta\bm{W}=\bm{B}\bm{A}$, where $\bm{B}\in\mathbb{R}^{d\times r}$ and $\bm{A}\in\mathbb{R}^{r\times k}$, and rank $r\ll\text{min}(d,k)$. During training, only $\bm{B}$ and $\bm{A}$ are updated, and other parameters of the pre-trained model are frozen. Since the trained matrices are significantly smaller than the updated weights, LoRA significantly reduces the number of fine-tuned parameters.

\textit{LayerNorm Tuning} (LNTuning) ~\citep{zhao2023tuninglayernormattentionefficient} is a PEFT method widely used for transforming large language models (LLMs) to Multi-Modal LLMs (MLLMs). 
It transitions a text understanding model to other modalities by adding new trainable parameters to targeted LayerNorm modules inside each attention block. During fine-tuning, only the newly introduced parameters are updated, and all the model's weights are kept frozen. Similarly to LoRA, LNTuning significantly reduces the number of trainable parameters.

\textbf{Differential Privacy (DP)} is a mathematical framework used to bound the possible privacy leakage a mechanism exhibits. In context of machine learning, the goal is to quantify and limit the extent of privacy risks associated with the data used to train a model. If presence or absence of a specific data point,~\eg an image, in the training set alters the behavior of the model significantly, then an adversary can learn about this data point from the model alone, which constitutes a privacy breach. To address this issue, DP provides guarantees on the level of impact a single data point can have on the model, giving an upper bound.  

Specifically, we have two datasets, $\mathcal{D}$ and $\mathcal{D'}$ differing by a single data point, and a mechanism $\mathcal{M}$, acting on $\mathcal{D}$ and $\mathcal{D'}$. Let $\mathcal{S}$ be a set of all possible outputs of $\mathcal{M}$. We say $\mathcal{M}$ is $(\epsilon,\delta)$-DP if
$Pr[\mathcal{M}(D) \in \mathcal{S}] \leq e^ \epsilon Pr[\mathcal{M}(D') \in \mathcal{S}] + \delta$. Intuitively, the maximum possible difference of outputs of $\mathcal{M}$ for any $\mathcal{D}$ and $\mathcal{D'}$ is bounded by an exponential factor dependent on $\epsilon$, and $\delta$ provides an optional safety margin.
The smaller the value of $\epsilon$, the stronger the privacy guarantees. 

The go-to algorithm to train DP models is DP-SGD~\cite{Abadi_2016}---a differentially private version of the regular SGD. DP-SGD first clips per-sample gradients, averages them, and adds Gaussian noise, obtaining privatized gradients. The parameters update at $i^{th}$ iteration is:

$\theta_{i+1} = \theta_i - \eta\left(\frac{1}{L}\sum_{k=1}^{L}\text{clip}(g(x_k))+\mathcal{N}(0,\sigma^2C^2I)\right)$,

where clip$(g(x_k))=g(x_k)/max(1,\frac{||g(x_k)||_2}{C})$, $\eta$ is the learning rate, $C$ is the clipping norm, $\sigma$ is the noise magnitude, $g(x_k)$ is the gradient for a single input data point $x_k$, $L$ is the lot size and $\theta$ are the parameters.

\textbf{DP Adaptations of Image Generative Models.}
The noise from DP-SGD harms the convergence. Augmentation multiplicity~\citep{de2022unlocking} addresses that by averaging per‐sample gradients over multiple views to increase signal to noise ratio.
For DMs, DPDM~\citep{dockhorn2023differentiallyprivatediffusionmodels} introduced noise multiplicity, averaging per‐sample gradients over multiple input noise draws, which was further extended to augmentation multiplicity by~\citep{ghalebikesabi2023differentially}. Since limiting the number of updated parameters also boosts performance, 
DP‐LDM~\citep{liu2024differentiallyprivatelatentdiffusion} shifts training into a lower‐dimensional latent space via a non‐private encoder. 

%% file: sec/03nondp_finetuning.tex
\section{Fine-Tuning of VAR}
\label{sec:nondp_finetuning}

\begin{figure*}[h!]
    \centering
    \includegraphics[width=1\linewidth]{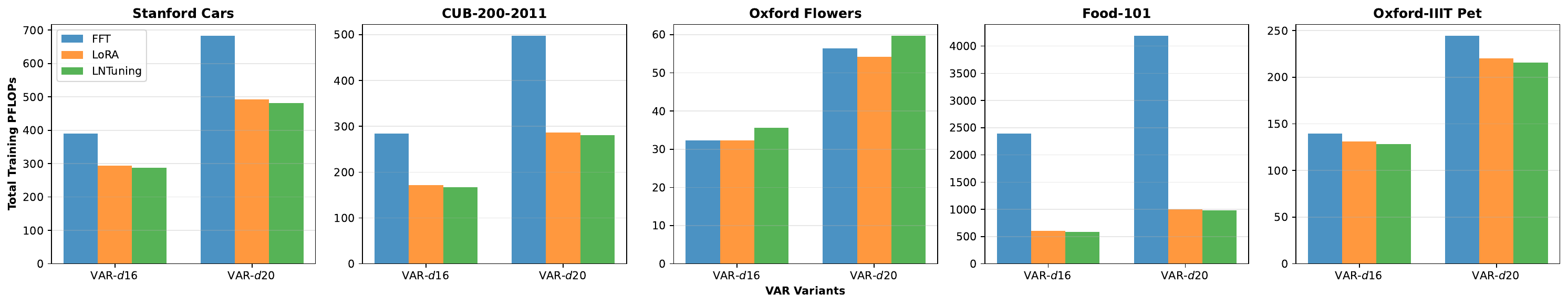}
    \vspace*{-20pt}
    \caption{\textbf{Training Compute Cost (PFLOPs) Comparison Across Datasets.}}
    \vspace{-10pt}
    \label{fig:compute-cost}
\end{figure*}

In the following we adapt VAR, a novel IAR, to specific downstream tasks via fine-tuning. We begin with a description of the models and datasets used, as well as the evaluation scheme. Then, we compare the performance of VAR to the one of SOTA adaptation method for DMs, DiffFit~\citep{xie2023difffitunlockingtransferabilitylarge}, and find that VAR outperforms its DM counterpart on five different datasets. Finally, we provide insights explaining observed discrepancy, and highlight key challenges we face during the implementation stage.

\subsection{Experimental Setup}
\label{sec:exp_set_nondp}

\begin{table*}[t]
    \scriptsize
    \centering
    \caption{\textbf{VAR fine-tuning outperforms DiffFit~\citep{xie2023difffitunlockingtransferabilitylarge}.} We compare FID ($ \downarrow $) on 5 downstream datasets between DiT-XL-2 and VAR-\textit{d}16 and VAR-\textit{d}20.}
    \label{tab:fid_inf_eps_comparision_same_size}
    \vspace{-10pt}
    \resizebox{\linewidth}{!}{%
        \begin{tabular}{cccccccc}
            \toprule
            \textbf{Model}      & \diagbox[width=3cm]{\textbf{Adaptation}}{\textbf{Dataset}} & \textbf{Food-101} & \textbf{CUB-200-2011} & \textbf{Oxford Flowers} & \textbf{Stanford Cars} & \textbf{Oxford-IIIT Pet} & \makecell{\textbf{Trainable}\\\textbf{Parameters}} \\
            \midrule
            DiT-XL-2            & FFT   & 10.46 & 5.68  & 21.05 & 9.79  & -    & 673.8M (100\%) \\
            DiT-XL-2            & LoRA           & 34.34 & 58.25 & 161.68& 75.35 & -    & 2.18M (0.32\%) \\
            DiT-XL-2            & DiffFit~\cite{xie2023difffitunlockingtransferabilitylarge}            & 6.96  & 5.48  & 20.18 & 9.90  & -    & 0.83M (0.12\%) \\
            \midrule

            VAR \textit{d}16     & FFT   & \textbf{6.11}  & \textbf{5.74}  & \textbf{12.08} & \textbf{7.42}  & 13.13 
& 309.6M (100\%)   
\\
            VAR \textit{d}16     & LoRA           & \textbf{6.94} & 7.84 & \textbf{13.18} & \textbf{8.87} & 13.70& 6.02M (1.91\%)   
\\
            VAR \textit{d}16     & LNTuning   & 8.01 & 8.15 & 22.82 & \textbf{9.27} & 14.28 & 100.7M (24.56\%) 
\\
            
            \midrule

            VAR \textit{d}20     & FFT   & \textbf{5.38}  & \textbf{5.58}  & \textbf{11.65} & \textbf{6.31}  & 12.81 & 599.7M (100\%)   
\\
            VAR \textit{d}20     & LoRA           & 6.97 & 6.29 & \textbf{11.16} & \textbf{9.42} & 12.97 & 9.42M (1.54\%)   
\\
            VAR \textit{d}20     & LNTuning   & 7.00& 6.07& \textbf{12.74} & \textbf{7.36} & 12.86& 196.7M (24.69\%) \\
            \bottomrule
        \end{tabular}%
    }
    \vspace{-15pt}
\end{table*}

\textbf{Models:} We perform evaluation on VAR-\textit{d}16 and VAR-\textit{d}20 pre-trained on ImageNet-1k~\citep{imagenet} to perform class-conditional image generation in 256x256 resolution, and are sourced from the repositories provided by their respective authors. We restrict ourselves to these two models, because they are of comparable sizes to DiT-XL-2~\citep{peebles2023scalable}. In~\cref{app:experiments_all_vars} we provide results for bigger models: VAR-\textit{d}24 and VAR-\textit{d}30. 

\textbf{Datasets:} As our downstream task, we focus on image generation in narrow domains, and fine-tune the models to perform well with small datasets. To this end, we use \textit{Food-101}~\citep{food101} with 101k total images of 101 different categories of food, \textit{CUB-200-2011}~\citep{cubbird} of 11,778 images of 200 bird classes, \textit{Oxford Flowers}~\citep{oxfordflowers} consisting of 1020 images of 102 species of flowers, \textit{Stanford Cars}~\citep{stanford_cars} with 16,185 images of 196 classes of cars, and \textit{Oxford-IIIT Pet Dataset}~\citep{parkhi2012cats} with 7,393 images of 37 different breeds of cats and dogs. More details about the datasets,~\eg size of the training and validation sets, can be found in~\cref{app:experimental_details}.

\textbf{Adaptations and Hyperparameters:} In our study we use LoRA and LNTuning as PEFT methods, and compare them to FFT. For LoRA, we use rank $r=16$, $\alpha=2r$, and LoRA dropout of 0. For LoRA fine-tuning we target self-attention modules,~\ie the query, key, and value matrices. Additionally, we fine-tune the projection layers of self-attention, and the Adaptive LayerNorm modules. For LNTuning, we update only the Adaptive LayerNorm modules. FFT updates all parameters of the model. We provide extended details in~\cref{app:experimental_details}.

\textbf{Performance Metrics:} We measure Fréchet Inception Distance (FID)~\citep{heusel2017fid} to quantify generation quality. For each dataset we generate as many samples as present in each class of the respective test sets, and compute FID between images in generated set and test set. Moreover, we compute the computation cost of fine-tuning, expressed in Peta Floating Point Operations (PFLOps). Full evaluation setup can be found in~\cref{app:experimental_details}.

\subsection{Empirical Results}
\label{sec:results_nondp}
\textbf{Compute Cost Comparison.}  
In~\cref{fig:compute-cost} we show that across all datasets and VAR variants, FFT incurs the highest compute cost, with the exception of Oxford Flowers, where the cost is similar for all adaptations. The biggest difference is visible for Food-101, where FFT requires around $4.5\times$ more compute than parameter-efficient counterparts. Interestingly, when the size of dataset increases (Food-101 has 101k samples, Oxford Flower only 1020), the difference in cost between FFT and PEFT also increases.

Among the two PEFT techniques, LNTuning exhibits a slightly smaller compute footprint than LoRA.

\textbf{Performance Comparison.}
While we observe significant differences in compute cost between the methods, we should take other factors into consideration,~\eg convergence speed, or the final generation quality. 

Importantly, LNTuning is the cheapest, while FFT performs best (lowest FID scores), according to the results in~\cref{tab:fid_inf_eps_comparision_same_size}. LoRA strikes the middle ground between these methods: it matches FFT’s FID performance, and is similarly demanding to LNTuning in terms of compute.

VAR fine-tuning achieves better performance than DiffFit, with FFT outperforming it across all models and datasets. It suggests that the adaptations of IARs might gain significance as the field progresses. We provide additional metrics in~\cref{app:experiments_all_vars}.

\begin{figure}[!htbp]
    \centering
    \vspace{-5pt}
    \includegraphics[width=1\linewidth]{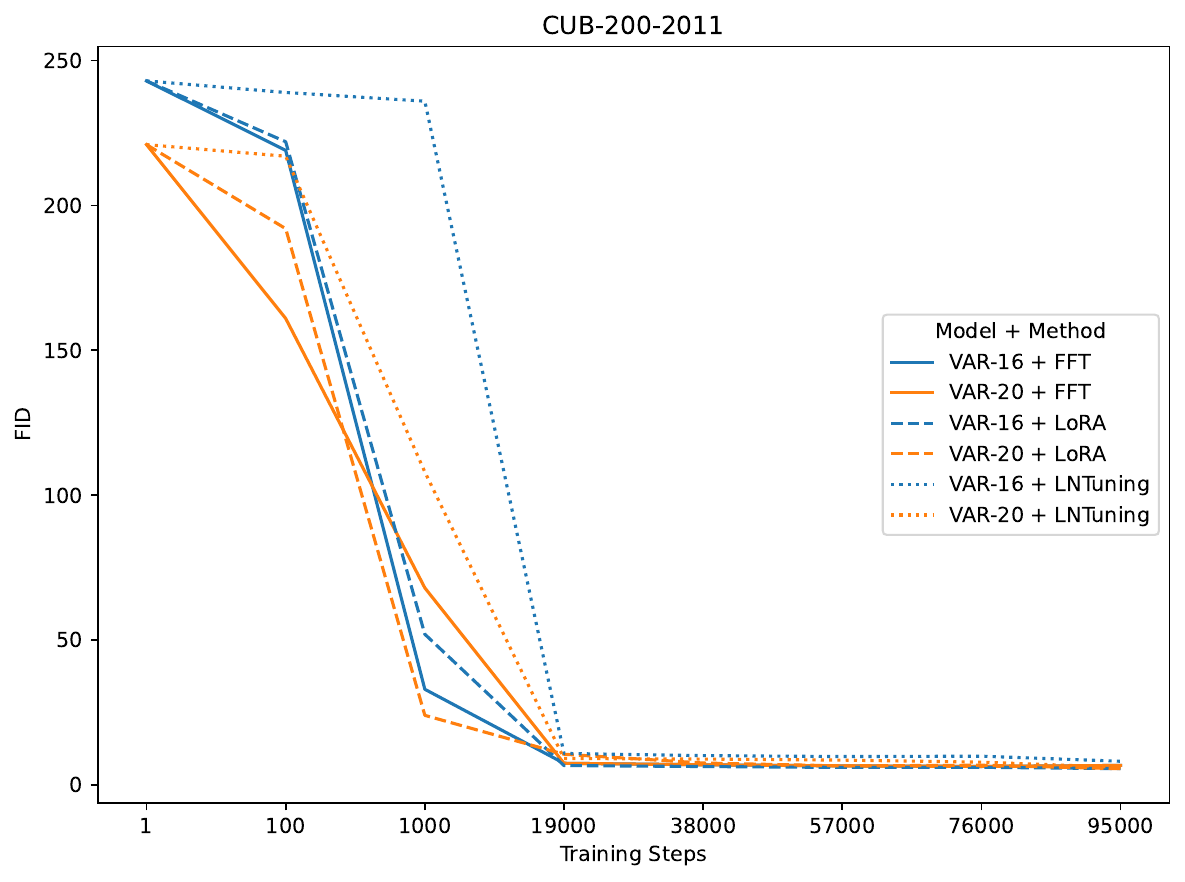}
    \caption{\textbf{VARs converge after small amount of training steps.} Dataset: CUB-200-2011.
    }
    \label{fig:var_convergence_cub200}
\end{figure}

\textbf{VARs Converge Quickly.}
During our experiments we notice that VAR needs very few update steps to reach its final performance. This contrasts with the behavior observed for DMs, which tend to require extended training, which involves stochastic input noise. We investigate the convergence behavior, and in~\cref{fig:var_convergence_cub200} we show that the models achieve their final FID after few thousand steps. Interestingly, LoRA fine-tuning converges similarly fast to FFT, while LNTuning needs more update steps. We provide more insights on why VAR is faster in~\cref{app:var_quick_convergence}.

\textbf{Implementation Details}
The original implementation of the attention operator in VAR requires patches (\cref{tab:patch_model}) to introduce LoRA adapters. We provide more details in~\cref{app:iar_implementation_details}.

%% file: sec/04dp_finetuning.tex
\section{Differentially-Private Adaptation for VAR}
\label{sec:dp_adaptations}

Next, we switch our focus to DP adaptations for VAR, whose goal is to preserve the privacy of potentially sensitive fine-tuning data. 
We show the impact of augmentation multiplicity, model's size, fine-tuning strategy, and the privacy budget $\epsilon$ on the generation quality.

\subsection{Experimental Setup}

Due to high computation cost of fine-tuning with DP, we only fine-tune the models on the Oxford Flowers dataset, and investigate LoRA and LNTuning adaptations. For augmentation multiplicity we use the default pre-processing image transformation of VAR, with parameter $k$ denoting how many views per sample we craft.

\subsection{Empirical Evaluation}
\label{sec:results_dp}

\textbf{Performance.}
Contrary to promising results in~\cref{sec:results_nondp}, when we fine-tune with DP, the models fail to converge. Our results in~\cref{tab:dp_epsilon_ablation} show that we need extremely high values of $\epsilon$ to obtain acceptable generation quality. Interestingly, we observe that augmentation multiplicity yields only modest improvements, with results for LoRA in~\cref{tab:lora_lntuning_k} indicating negligible gains at $k=128$. Notably, the compute cost for $k=128$ is around 128 times greater for $k=1$, and increasing $k$ further might be prohibitively expensive. LoRA appears to outperform LNTuning, according to~\cref{tab:lora_lntuning_k}, which might be due to the lower number of trainable parameters for LoRA (see~\cref{tab:fid_inf_eps_comparision_same_size}).

\begin{table}[H]
    \scriptsize
    \centering
    \caption{\textbf{FID ($ \downarrow $)} of VAR models fine-tuned using LoRA and DP with $\epsilon=10$ on Oxford Flowers Dataset.}
    \resizebox{\linewidth}{!}{
    \label{tab:lora_lntuning_k}
    \begin{tabular}{cccc}
            \toprule
            & \multicolumn{2}{c}{LoRA} & LNTuning \\
            \textbf{Model} & $k=1$ & $k=128$ & $k=1$ \\
            \midrule
            VAR-\textit{d}16 & 69.92 & 63.24 & 106.32  \\
            VAR-\textit{d}20 & 68.92 & 59.29 & 98.2  \\
            \bottomrule
    \end{tabular}
    }
\end{table}

\begin{table}[H]
    \scriptsize
    \centering
    \caption{
    \textbf{FID ($ \downarrow $)} of VAR models fine-tuned using LoRA of varying $\epsilon$ on Oxford Flowers Dataset with augmentation multiplicity parameter $k=32$.}
    \label{tab:dp_epsilon_ablation}
    \renewcommand{\arraystretch}{1.5}
    \resizebox{\linewidth}{!}{%
        \begin{tabular}{cccccccc}
            \toprule
            \textbf{Model} & $\epsilon=1$ & $\epsilon=10$ & $\epsilon=20$ & $\epsilon=50$ & $\epsilon=100$ & $\epsilon=500$ & $\epsilon=1000$ \\
            \midrule
            VAR-\textit{d}16 & 196.52& 60.24& 52.10& 46.36& 41.63& 35.70& 35.36\\
            VAR-\textit{d}20 & 160.33& 63.38& 53.73& 47.09& 43.35& 37.35& 35.06\\
            \bottomrule
        \end{tabular}%
    }
\end{table}

\textbf{Implementation.}
Similarly as for LoRA adaptations, tailoring the original VAR implementation for DP fine-tuning requires patches. We identify issues with model-specific buffers and a non-standard forward function. We provide more details in~\cref{app:iar_implementation_details}.

%% file: sec/05conclusions.tex
\section{Conclusions}
\label{sec:conclusions}
With the success of our adaptations of the VAR model, we expect the broader adoption of IARs tailored specific domains for image generation. Our work is the first step in the direction of IAR-specific model adaptations, and we already observe that fine-tuned VAR performs better than the SOTA adaptation strategy for DMs. Importantly, DP adaptations remain ineffective, which indicates that further research in that direction is required to enable privacy-preserving adaptations on sensitive data.
\clearpage

%% file: sec/10ack.tex
\section*{Acknowledgments}

This project was funded by the Deutsche Forschungsgemeinschaft (DFG, German Research Foundation), Project number 550224287, and by the Helmholtz Impulse and Networking Fund as part of the project ``Effective Privacy-Preserving Adaptations of Foundation Models for Medical Tasks'', reference number ZT-I-PF-5-227.

%% file: sec/20appendix.tex
\appendix

\section{Additional Related Work}
\label{sec:app-related-work}

\subsection{Properties of Differential Privacy}
DP has the properties of group privacy, composability, and robustness to auxiliary information. Group privacy ensures graceful degradation of privacy guarantees when datasets have correlated inputs. If a dataset has $k$ such points, then an ($\epsilon$, $\delta$) algorithm run on such a dataset yields ($k\epsilon$, $ke^{(k-1)\epsilon}\delta$)-DP~\cite{dwork2006}. Composability ensures that if in a mechanism, each of the components are differentially private then the mechanism is differentially private too. Robustness to auxiliary information implies that even with the auxiliary knowledge, DP guarantees hold against an adversary.
A common approach to implement DP guarantees is to add noise, which magnitude depends on the sensitivity of the mechanism. Sensitivity of a mechanism is defined as the maximum distance between outputs of the mechanism for any two adjacent datasets. Therefore a mechanism $f$ can be made differentially private using Gaussian noise as follows:
\begin{center}
    $\mathcal{M}(d) = f(d) + \mathcal{N}(0,\Delta f^2\sigma^2)$
\end{center}

where $\Delta f$ is the sensitivity of mechanism $f$ and $\sigma$ is the noise scale. Such a mechanism is called a \textit{Gaussian mechanism}.

\subsection{Differentially Private Diffusion Models}

\paragraph{DPDM.}~\citet{dockhorn2023differentiallyprivatediffusionmodels} begin from a fact that the gradients at each update step are prone to high variance, because the training objective relies on the forward process, which adds noise with a varying magnitude. By default, this issue is alleviated by training for many iterations at a small batch size. However, DP-SGD computes privacy loss on a per-update basis, which means such an approach is not feasible.

The non-private DM loss can be expressed as 

\begin{equation}
    l_i = \lambda(\sigma_i)||z_\theta(x_i +n_i,\sigma_i) - x_i ||_2^2
\end{equation}
where, $l_i$ is the loss, $x_i + n_i$ is the noisy input passed to the DM, $\lambda$ is a weighting function and $\sigma_i$ is the noise scale for the forward process. DPDM tackles the problem of $l_i$ using the \textit{noise multiplicity}, which replaces the DM objective function over one noisy sample, with the average of $K$ noisy samples for each data point. Effectively, the new loss function can be represented as
\begin{equation}
    \tilde{l_i} = \frac{1}{K}\sum_{k=1}^{K} \lambda(\sigma_{ik})||D_\theta(x_i +n_{ik},\sigma_{ik}) - x_i ||_2^2
\end{equation}
The results show a significant improvement of the performance after incorporating this mechanism to the private training. The gain from the noise multiplicity plateaus around $K=32$.

In addition to the noise multiplicity, authors use exponential moving average to update the weights of the DM after each training step, and very large batch sizes (4096 for MNIST~\cite{lecun1998mnist}. They also note that biasing the forward process towards higher timesteps is beneficial for the training.

\citet{ghalebikesabi2023differentially} show that by sampling an augmentation in addition to the timestep at each repetition in the noise multiplicity mechanism yields an improvement. They dub that approach \textit{augmentation multiplicity}. They find that $K=128$ yields the best results, with batch size of $16,384$. Similarly to DPDM, they sample the timesteps from a distribution different than the default $\mathcal{U}[0,T]$, $t \sim \sum_{i=1}^K w_i \mathcal{U}[l_i, u_i], \quad \text{where } \sum_{i=1}^K w_i = 1, \, 0 \leq l_0, u_K \leq T, \, \text{and } u_{k-1} \leq l_k \, \text{for } k \in \{2, \dots, K\}$. The main difference between their work and DPDM is that they first pre-train the DM using "public" dataset without DP, and only then fine-tune on the private dataset, this time with DP.

\paragraph{DP-LDM.}~\citet{liu2024differentiallyprivatelatentdiffusion} focus on the impact of the size of the DM on the properties of DP-SGD. Intuitively, the bigger the size of the model, the bigger the norm of the gradient, in effect the bigger the noise added by the DP-SGD mechanism. Thus, a natural idea is to limit the size of the DM. To this end, authors utilize Latent Diffusion Models (LDMs)~\citep{rombach2022high}, which consist of of a VAE and a DM. VAEs are made of an encoder, which takes a high-dimensional image in the pixel space as an input, and produces a lower-dimensional latent representation, and a decoder that converts the latent back to the pixel space. Effectively, the DM is trained in the latent space, and in turn contains less parameters than a pixel-space alternative. DP-LDM first trains the VAE (without DP), and, similarly to~\cite{ghalebikesabi2023differentially}, pre-trains the DM on "public" dataset. To further benefit from the smaller gradient norms, they fine-tune with DP only the (cross-)attention modules of the DM, which amounts to $10\%$ of the parameters.

\paragraph{DP-LoRA.}~\citet{tsai2024differentiallyprivatefinetuningdiffusion} improve over DP-LDM by utilizing a PEFT method--LoRA--to limit the gradient norms by reducing the number of trained parameters of the DM even further. LoRA adapters are applied to the QKV matrices of the attention blocks of the LDM, and the output projection layer following them.

\section{Extended Experimental Details}
\label{app:experimental_details}

\paragraph{Additional information about the datasets.} All variants of VAR are pre-trained on ImageNet 256x256~\citep{imagenet}, we finetune them using the datasets mentioned in~\cref{tab:dataset_stats}. We use the same datasets for regular finetuning and DP-finetuning as well. All experiments in paper leverage the entire train split for finetuning. We use 100\% of the test split when benchmarking the results for the experiments.

\FloatBarrier
\begin{table}[!htbp]
    \centering
    \caption{Different datasets used across our experiments with VARs.}
    \label{tab:dataset_stats}
    \renewcommand{\arraystretch}{1}
    \resizebox{\linewidth}{!}{%
        \begin{tabular}{cccc}
            \toprule
             Dataset  &Num. Classes&Training Set Size&Testing Set Size\\
            \midrule
             CUB-200-2011~\citep{cubbird} &200&5994&5794\\ 
             Food-101~\citep{food101}&101&75750&25250\\
             Oxford Flowers~\citep{oxfordflowers}&102&1020&6149\\
             Oxford-IIIT Pet~\citep{parkhi2012cats}&37&3680&3669\\
             Stanford Cars~\citep{stanford_cars}&196&8144 &8041 \\
             \bottomrule
        \end{tabular}
    }
\end{table}
\FloatBarrier

\paragraph{Hyperparameters used in the experiments (Non-Private finetuning).} We apply identical hyperparameter settings across all variants of each model: VAR-\textit{d}\{16, 20, 24, 30\}. Larger variants typically require fewer epochs and allow larger batch sizes. The hyperparameters for VARs are given in~\cref{tab:eps-inf-hyperpara-var-fft},~\cref{tab:eps-inf-hyperpara-var-lora} and~\cref{tab:eps-inf-hyperpara-var-lnt}.

\FloatBarrier
\begin{table}[!htbp]
  \centering
  \caption{Hyperparameters for FFT.}
  \label{tab:eps-inf-hyperpara-var-fft}
  \renewcommand{\arraystretch}{1.2}
  \begin{tabular}{lccc}
    \toprule
    Dataset             & Learning Rate          & Batch Size & Epochs \\
    \midrule
    CUB-200-2011        & \(1\times10^{-4}\)     & 512        & 50     \\
    Food-101            & \(1\times10^{-4}\)     & 256        & 30\\
    Oxford Flowers      & \(1\times10^{-4}\)     & 256        & 24     \\
    Oxford-IIIT Pet     & \(1\times10^{-4}\)     & 256        & 40     \\
    Stanford Cars       & \(1\times10^{-4}\)     & 256        & 50     \\
    \bottomrule
  \end{tabular}
\end{table}
\FloatBarrier

\FloatBarrier
\begin{table}[!htbp]
  \centering
  \caption{Hyperparameters for LoRA.}
  \label{tab:eps-inf-hyperpara-var-lora}
  \renewcommand{\arraystretch}{1.2}
  \begin{tabular}{lcccc}
    \toprule
    Dataset             & Learning Rate          & Batch Size & LoRA Rank & Epochs \\
    \midrule
    CUB-200-2011        & \(1\times10^{-3}\)     & 512        & 16        & 40     \\
    Food-101            & \(5\times10^{-4}\)     & 256        & 16        & 6\\
    Oxford Flowers      & \(5\times10^{-4}\)     & 256        & 16        & 40\\
    Oxford-IIIT Pet     & \(5\times10^{-4}\)     & 256        & 16        & 50\\
    Stanford Cars       & \(5\times10^{-4}\)     & 256        & 16        & 50     \\
    \bottomrule
  \end{tabular}
  
\end{table}
\FloatBarrier

\FloatBarrier
\begin{table}[!htbp]
  \centering
  \caption{Hyperparameters for LNTuning.}
  \label{tab:eps-inf-hyperpara-var-lnt}
  \renewcommand{\arraystretch}{1.2}
  \begin{tabular}{lccc}
    \toprule
    Dataset             & Learning Rate          & Batch Size & Epochs \\
    \midrule
    CUB-200-2011        & \(1\times10^{-3}\)     & 512        & 40     \\
    Food-101            & \(5\times10^{-4}\)     & 256        & 6\\
    Oxford Flowers      & \(5\times10^{-4}\)     & 256        & 45\\
    Oxford-IIIT Pet     & \(5\times10^{-4}\)     & 256        & 50\\
    Stanford Cars       & \(5\times10^{-4}\)     & 256        & 50     \\
    \bottomrule
  \end{tabular}
\end{table}
\FloatBarrier

\paragraph{Hyperparameters for private finetuning.}  
Differentially-private fine-tuning typically requires more epochs to offset the utility loss from added noise, so we increase the epoch count accordingly. All DP-finetuning experiments are conducted exclusively on the Oxford Flowers dataset. We set the total batch size \(BS_{\mathrm{total}}\) from the sampling rate \(q\) and total examples \(N\) as:
\[
BS_{\mathrm{total}} \;=\;\lfloor q \times N\rfloor,
\]
and in a multi-GPU (DDP) run with \(G\) GPUs the effective batch size per optimizer step is:
\[
BS_{\mathrm{device}} \;=\; \frac{BS_{\mathrm{total}}}{G}.
\]

\FloatBarrier
\begin{table}[!htbp]
  \centering
  \caption{Hyperparameters for DP-finetuning (DP-$\varepsilon-10$) on Oxford Flowers for VAR.}
  \label{tab:dp-hyperpara-flowers-var}
  \renewcommand{\arraystretch}{1.2}
  \begin{tabular}{llc}
    \toprule
    Method        & Hyperparameter            & Value        \\
    \midrule
    \multirow{6}{*}{FFT}     
                  & Learning Rate             & \(1 \times 10^{-4}\)\\
                  & Sample Rate (\(q\))       & 0.251\\
                  & Delta (\(\delta\))        & \( \frac{1}{N} \) \\
                  & Epochs                    & 180 \\
                  & Max Grad Norm             & 0.1\\
                  & Augmentation Multiplicity (\(k\))                    & 128\\
    \midrule
    \multirow{7}{*}{LoRA}    
                  & Learning Rate             & \(5 \times 10^{-4}\)\\
                  & Sample Rate (\(q\))       & 0.251\\
                  & Delta (\(\delta\))        & \( \frac{1}{N} \) \\
 & LoRA Rank                 &16\\
                  & Epochs                    & 80\\
                  & Max Grad Norm             & 0.5\\
                  & Augmentation Multiplicity (\(k\))                     & 128\\
    \midrule
    \multirow{6}{*}{LNTuning} 
                  & Learning Rate             & \(5 \times 10^{-4}\)\\
                  & Sample Rate (\(q\))       & 0.251\\
                  & Delta (\(\delta\))        & \( \frac{1}{N} \) \\
                  & Epochs                    & 100\\
                  & Max Grad Norm             & 0.5\\
                  & Augmentation Multiplicity (\(k\))                     & 128\\
    \bottomrule
  \end{tabular}
\end{table}
\FloatBarrier

\paragraph{Profiling}

We use built-in profiler module inside PyTorch~\citep{paszke2019pytorch} for computing the total training cost for each of our experiment. The designated profiling script utilizes the \textit{PyTorch Profiler} to aggregate the total number of FLOps for one single training step, this involves computing the total cost for 2 events inside the training step, forward pass and the backward pass. Finally, we calculate the cost for the whole run, based on the effective batch size and number of epochs multiplied by the cost of each step to obtain the final compute cost in PFLOps. 

\paragraph{Compute Cost Profiling for DP-experiments}Unlike the non-private baselines, each differentially-private (DP) experiment employs \emph{Poisson subsampling}: at every update each example $i\!\in\!\{1,\dots,N\}$ is drawn independently with probability $q\!=\!\tfrac{\text{batch\_size}}{N}$.  
Because $|B_t|\sim\operatorname{Binomial}(N,q)$, the FLOps for each update inherit a coefficient of variation $\tfrac{1}{\sqrt{qN}}$; hence multiplying a single-step profile by the expected step count yields only a coarse approximation of the total compute budget. Profiling only the \emph{first} update and multiplying by an \emph{expected} step count therefore produces a point estimate that is systematically biased and may lead to incorrect calculations. To avoid reporting a misleadingly precise figure, we refrain from quoting aggregate PFLOp totals for DP experiments and instead restrict compute-cost analysis to the deterministic, fixed-batch baselines.

\section{Extended Experimental Results for Non-Private Experiments}\label{app:experiments_all_vars}

\textbf{Extended Metrics for Performance.} Following~\citep{sajjadi2018towards,kynkaanniemi2019improved}, we evaluate VAR-\textit{d}16 and VAR-\textit{d}20 to estimate a local $k$–NN manifold for both the real and the generated distributions and report \textbf{Precision} and \textbf{Recall} in~\cref{tab:precision_and_recall_results}. Precision ($\uparrow$) is the fraction of generated samples that fall inside the real manifold and thus quantifies \emph{sample fidelity}. Recall ($\uparrow$) is the fraction of real samples that fall inside the generated manifold and therefore measures \emph{coverage/diversity}.

\FloatBarrier
\begin{table}[!htbp]
  \centering
  \caption{Extended Precision (\textbf{P}) and Recall (\textbf{R}) of VAR-\textit{d}16 and VAR-\textit{d}20 models across five downstream datasets.}
  \label{tab:precision_and_recall_results}
  \resizebox{\linewidth}{!}{%
    \begin{tabular}{cc *{5}{cc} c}
      \toprule
      \textbf{Model} & \textbf{Adaptation}
        & \multicolumn{2}{c}{\textbf{Food-101}}
        & \multicolumn{2}{c}{\textbf{CUB-200-2011}}
        & \multicolumn{2}{c}{\textbf{Oxford Flowers}}
        & \multicolumn{2}{c}{\textbf{Stanford Cars}}
        & \multicolumn{2}{c}{\textbf{Oxford-IIIT Pet}}
        & \makecell{\textbf{Trainable}\\\textbf{Parameters}} \\
      \cmidrule(lr){3-4}\cmidrule(lr){5-6}\cmidrule(lr){7-8}\cmidrule(lr){9-10}\cmidrule(lr){11-12}
         & 
        & \textbf{P} & \textbf{R}
        & \textbf{P} & \textbf{R}
        & \textbf{P} & \textbf{R}
        & \textbf{P} & \textbf{R}
        & \textbf{P} & \textbf{R}
        & \\ 
      \midrule
      VAR-\textit{d}16 & FFT      & 73.48\% & 7.65\% & 61.99\% & 58.40\% & 59.45\% & 26.57\% & 58.40\% & 40.03\% & 68.49\% & 2.67\% & 309.6M (100\%)   \\
      VAR-\textit{d}16 & LoRA     & 66.89\% & 9.70\% & 63.15\% & 58.16\% & 61.68\% & 26.83\% & 47.24\% & 31.20\% & 69.77\% & 4.52\% & 6.02M (1.91\%) \\
      VAR-\textit{d}16 & LNTuning & 67.67\% & 8.43\% & 48.15\% & 62.44\% & 42.57\% & 34.85\% & 43.72\% & 41.38\% & 57.23\% & 4.33\% & 100.7M (24.56\%) \\
      \midrule
      VAR-\textit{d}20 & FFT      & 73.63\% & 8.07\% & 74.74\% & 47.99\% & 64.17\% & 24.84\% & 59.01 & 39.68\% & 69.28\% & 2.28\% & 599.7M (100\%)   \\
      VAR-\textit{d}20 & LoRA     & 68.90\% & 9.36\% & 68.24\% & 54.26\% & 59.64\% & 32.18\% & 58.97\% & 39.49\% & 68.08\% & 3.08\% & 9.42M (1.54\%) \\
      VAR-\textit{d}20 & LNTuning & 69.24\% & 10.03\% & 61.23\% & 58.08\% & 59.27\% & 31.71\% & 51.11\% & 44.58\% & 65.46\% & 3.40\% & 196.7M (24.69\%) \\
      \bottomrule
    \end{tabular}
  }
\end{table}
\FloatBarrier

Recall values dip sharply for \emph{Food-101} (max.\ $\leq10\%$) and \emph{Oxford-IIIT Pet} (all methods $<\!5\%$). Both datasets are (i) \emph{fine-grained}, demanding the model to cover
hundreds of subtly different classes, and (ii) \emph{out-of-domain} for the ImageNet-trained encoder used by the metric, which clusters visually similar species into overly tight neighbourhoods. Under these conditions the k-NN test rejects many legitimate but stylistically
rare samples, driving recall down even for FFT.

\begin{figure}[!htbp]
    \centering
    \includegraphics[width=1\linewidth]{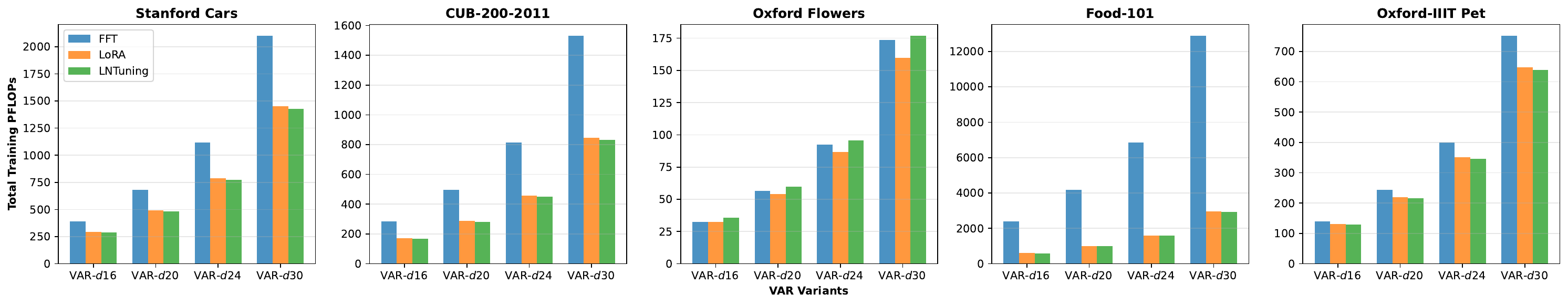}
    \caption{Extended Training Compute Cost (PFLOPs) Comparison Across Datasets.}
    \label{fig:extended-compute-cost}
\end{figure}

\textbf{FLOPs Compute Cost Analysis.} We show extended FLOPs compute‐cost analysis for VAR-\textit{d}\{16, 20, 24, 30\} in~\cref{fig:extended-compute-cost}. These results expand upon our primary evaluation, which originally included only the VAR-\textit{d}16 and VAR-\textit{d}20 models. By adding the larger VAR-\textit{d}24 and VAR-\textit{d}30 variants, we provide a full picture of how scaling the autoregressive backbone affects total PFLOPs for each adaptation method.

The plots clearly show that LoRA and LNTuning deliver greater compute savings as model size grows. While for FFT the PFLOPs increase steeply from VAR-\textit{d}16 through VAR-\textit{d}30, both PEFT methods remain comparatively flat—especially on large datasets like Food-101 and Stanford Cars. All compute costs were calculated using the identical hyperparameters specified in ~\cref{tab:eps-inf-hyperpara-var-fft,tab:eps-inf-hyperpara-var-lora,tab:eps-inf-hyperpara-var-lnt}, reinforcing that PEFT is the most scalable strategy under a fixed compute budget.  

On the Oxford Flowers dataset, both LoRA and LNTuning were scheduled for more epochs than FFT (\eg 30 vs. 24), yet their total PFLOPs remain lower. This occurs because the per‐step overhead of PEFT methods is substantially smaller than that of FFT, so even with an extended training schedule, the aggregate compute cost stays below FFT’s. Consequently, PEFT retains its efficiency advantage on Flowers despite requiring more epochs.

\section{Ablation Study}

\textbf{LoRA outperforms LNTuning in DP experiments.} 

We check what happens with the performance of VAR when we select different parameters to fine-tune with DP-SGD. To this end, we run a controlled ablation on VAR-\textit{d}16 using the Oxford Flowers dataset. 
We use the same hyperparameters as in~\cref{app:experimental_details}. We keep LoRA rank fixed at 16 and vary which weight blocks receive the low-rank adapters: (i) LoRA-A updates only the attention projections \(W_{\mathrm{qkv}}\) and the output projections; (ii) LoRA-M updates only the MLP projections $fc_1$ and $fc_2$; (iii) LoRA-AM + LN combines (i) + (ii) and additionally includes all LayerNorm weights; Finally, (iv) we include LNTuning as the baseline that trains an adapter for LayerNorm only. Effectively, we ablate over the capacity (number of parameters to fine-tune) and layers to fine-tune.~\cref{tab:dp_peft_abla} reports FID and number of trainable parameters for each variant.

\begin{table}[!htbp]
    \centering
\caption{Effect of selectively finetuning specific components of VAR-\textit{d}16 with DP $\varepsilon=10$ on Oxford Flowers Dataset with augmentation multiplicity $k = 128$ to compare FID ($ \downarrow $) scores.}
\label{tab:dp_peft_abla}
    \begin{tabular}{cccc}\toprule
         Adaptation Variant&  FID ($ \downarrow $)&  Trainable Parameters& Notes\\\midrule
         LoRA-A&  78.35&  1.57M (0.50\%)& Finetuning only Attention Layers\\
         LoRA-M&  81.78&  2.62M (0.84\%)& Finetuning only MLP Layers.\\
         LoRA-AM + LN&  63.24&  6.02M (1.91\%)& Main Reported LoRA  Method\\
         LNTuning&  116.64&  100.7M (24.56\%)& Main Reported LNTuning Method\\ \bottomrule
    \end{tabular}
\end{table}

Our results clearly show that a proper selection of layers to fine-tune is more crucial than the number of fine-tuned parameters. This happens due to the specifics of privacy training: information from gradients costs privacy, how the signal is allocated in the model affects its final performance.
Low-rank adapters placed in the attention and MLP projections capture task-specific directions that resists DP perturbations, whereas LNTuning lacks the expressive power to counteract the injected noise. Consequently, our main LoRA-AM + LN configuration delivers the best privacy–utility trade-off, outperforming LNTuning in FID while updating fewer than 2\% of the weights.

\section{Why VAR converges faster?}
\label{app:var_quick_convergence}

Results in~\cref{fig:var_convergence_cub200} indicate that VAR converges surprisingly fast to its downstream task when fine-tuned. We argue that this behavior stems from VAR's training objective. Specifically, VAR is trained with the standard cross-entropy loss between predicted and ground-truth tokens. With this objective, the gradient signal remains strong even early in training. Instead, DMs minimize a denoising score-matching loss, which tries to reconstruct clean latents from various noise levels.  The signal in this objective is diluted across $T$ noise scales; gradients for very noisy time-steps are dominated by injected Gaussian noise, slowing effective learning until the model acquires a good global estimate of the score function. 

\clearpage

\section{Implementation Details of Adaptations and DP for VAR}
\label{app:iar_implementation_details}

IARs have model architectures identical to the popular transformer models like GPT-2 \cite{radford2019language}. The IARs' state-of-the-art next-token prediction mechanism excels at generating high-resolution images significantly faster than the average diffusion model. During our research, we implemented our own finetuning pipeline for Non-private and Private finetuning of IARs. The code base allows finetuning pre-trained IARs on private datasets with DP guarantee and compare that to our Non-Private baseline. Our implementation presents private IARs retaining high quality images on several datasets.

\subsection{Overcoming Technical Barriers in Differentially Private IARs}

Implementing DP guarantees is not straightforward with many IARs as their functioning is different than how DMs generally work. Each implementation, Private or Non-Private requires some modification to allow different finetuning adaptations function properly. Considering one of the class conditional IAR like VAR \cite{VAR} registers buffers which are not supported by Opacus due to their nature of being a non-trainable parameter as well as the \textit{SelfAttention} mechanism causes empty gradient flow when implementing LoRA finetuning. To overcome these constraints we patch the internal code and calling the modified classes from the updated module while retaining the same performance and efficiency as before.

\FloatBarrier
\begin{table}[!htbp]
  \centering
  \caption{Various patching mechanisms implemented to successfully perform private and non-private finetuning with VARs.}
  \label{tab:patch_model}
  \renewcommand{\arraystretch}{1.2}
  \resizebox{0.75\linewidth}{!}{%
    \begin{tabular}{lccc}
      \toprule
       \textbf{Adaptation}            & \textbf{Patch Buffers} & \textbf{Override Forward Function} & \textbf{Patch \textit{SelfAttention}}  \\
      \midrule
       Full Fine-tuning               & \xmark & \xmark & \xmark  \\
         LoRA Fine-tuning               & \xmark & \xmark & \cmark  \\
         LayerNorm Fine-tuning          & \xmark & \xmark & \xmark  \\
         DP-Full Fine-tuning            & \cmark & \cmark & \xmark  \\
         DP-LoRA Fine-tuning            & \cmark & \cmark & \cmark  \\
         DP-LayerNorm Fine-tuning       & \cmark & \cmark & \xmark  \\
      \bottomrule
    \end{tabular}%
  }
\end{table}
\FloatBarrier

In this paper, we illustrate our approach to managing buffers, as they are typically found in most of the IAR implementations discussed.

\subsection{Patching Buffers to handle DP-finetuning}

IARs like VAR ~\citep{VAR} model register three important buffers that are crucial for its functioning. These buffers primarily handle embeddings, attention mask and bias making them highly important in order for VARs to function. The first buffer \textit{lvl\_1L}, is a level index tensor where each element contains the pyramid level index for the corresponding position in the sequence based on the patch numbers. This helps the model distinguish between tokens from different pyramid levels in the hierarchical structure. Another buffer registered in the VAR model is \textit{attn\_bias\_for\_masking} which is an Auto-regressive attention mask to ensure that tokens can only attend from the same or earlier pyramid levels. Lastly, the \textit{zero\_k\_bias} is a zero-filled tensor with shape \textit{embed\_dim} that serves as a placeholder bias for the key projection component. While Query and Value have learnable biases and are initialized as parameters, the \textit{zero\_k\_bias} is always passed as a fixed zero value tensor.

Unlike parameters, buffers are not updated at all and are not meant to have gradients computed either in batches or per-sample. Since the DP-SGD algorithm solely relies on computing gradients for each individual sample, Opacus does not handle the non-trainable buffer and causes the trainer to break immediately. Generally, buffers are removed in standard DP-SGD practices to avoid these errors although removing buffers from the VAR model results in unstable and garbled image generation since tokens are not positioned consistently since \textit{lvl\_1L} buffer does not exist and the attention bias does not exist to ensure correct level of scaled resolution.

We overcome buffer removal by not registering them at all in the first place and then passing them as functions with property decorators so Opacus doesn't create any sort of conflict with those.






\begin{figure}[htbp]
  \centering

  \begin{minipage}{0.9\textwidth}
    \begin{lstlisting}[language=Python]
attn_bias_for_masking = torch.where(
    d >= dT, 0., -torch.inf
).reshape(1, 1, self.L, self.L)

self.register_buffer(
    'attn_bias_for_masking',
    attn_bias_for_masking.contiguous()
)
    \end{lstlisting}
    \captionof{subfigure}{Original buffer implementation}
    \vspace{0.5em}
  \end{minipage}

  \begin{minipage}{0.9\textwidth}
    \begin{lstlisting}[language=Python]
@property
def attn_bias_for_masking(self):
    d = torch.cat([
        torch.full((pn * pn,), i,
                   device=self.pos_1LC.device)
        for i, pn in enumerate(self.patch_nums)
    ]).view(1, self.L, 1)
    dT = d.transpose(1, 2)
    attn_bias = torch.where(d >= dT,
                            0.0, -torch.inf
                   ).reshape(1, 1, self.L, self.L)
    return attn_bias.contiguous()
    \end{lstlisting}
    \captionof{subfigure}{Patched buffer implementation}
  \end{minipage}

  \caption{Buffer patching enables skipping \textit{register\_buffer} calls to avoid parameter incompatibility with Opacus.}
  \label{fig:buffer-patching}
\end{figure}

This way we do not have to register any buffers in the first place for them to conflict with the finetuning process while every time the model calls these newly patched functions they are computed on the fly and act like model attributes. This approach is lightweight and flexible with different types of buffers which are essential while making sure the arbitrary computation does not slow down the finetuning process.

\subsection{Addressing Gradient Flow Issues in \textit{SelfAttention} Layers for LoRA \& DP-LoRA in VARs}

The \textit{SelfAttention} component of the the VAR model contains two base layers we target when passing LoRA configuration to the model. One of the layers \textit{mat\_qkv} was computed by combining several operations in a single line, performing a linear transformation, adding the bias and reshaping to match the tensor shape. When implementing LoRA finetuning on this layer caused empty gradient propagation meaning the gradients were not flowing through this layer at all due to the condensed approach of \textit{qkv} computation. The problem occurs because Opacus needs to track and clip gradients at all times and non-initialized gradients causing to break the gradient flow. Simply meaning when using parameter-efficient fine-tuning methods like LoRA that rely on precise gradient computation for their low-rank updates the gradient computation becomes opaque to the DP mechanism.

We resolved this issue by decomposing the self-attention operation into its constituent parts, which benefits both standard LoRA and differentially private DP-LoRA implementations. Our patched approach follows three distinct steps: first applying the linear transformation through the LoRA-augmented layer \textit{self.mat\_qkv(x),} then explicitly adding biases as a separate operation, and finally reshaping the output tensor to avoid any tensor shape mismatches.





\begin{figure}[htbp]
  \centering

  \begin{minipage}{\textwidth}
    \begin{lstlisting}[language=Python]
qkv = F.linear(
    input=x,
    weight=self.mat_qkv.weight,
    bias=torch.cat(
        (self.q_bias, self.zero_k_bias, self.v_bias)
    )
).view(B, L, 3, self.num_heads, self.head_dim)
    \end{lstlisting}
    \captionof{subfigure}{Original \textit{qkv} computation method}
    \vspace{0.5em}
  \end{minipage}

  \begin{minipage}{\textwidth}
    \begin{lstlisting}[language=Python]
qkv_raw = self.mat_qkv(x)
qkv = qkv_raw + torch.cat(
    (self.q_bias, self.zero_k_bias, self.v_bias)
).view(1, 1, -1)
qkv = qkv.view(B, L, 3, self.num_heads, self.head_dim)
    \end{lstlisting}
    \captionof{subfigure}{Patched \textit{qkv} implementation}
  \end{minipage}

  \caption{Patching \textit{qkv} computation to add support for LoRA fine-tuning in VAR models \cite{VAR}.}
  \label{fig:qkv-patching}
\end{figure}

For standard LoRA, this separation ensures that the low-rank adaptation matrices properly participate in the computation graph, allowing gradients to flow correctly through both the base parameters and the LoRA adaptation matrices during backpropagation. For DP-LoRA specifically, this explicit separation is even more crucial, as it enables Opacus to correctly track, compute, and clip gradients at each step of the computation.

The modification maintains the mathematical equivalence of the operation while significantly improving the training dynamics for both approaches.

\subsection{Override Forward Function}

VAR models define a custom \textit{forward} method with two separate inputs \textit{label\_B} and \textit{x\_BLCv\_wo\_first\_l}, plus on‐the‐fly randomness for conditional dropout. Opacus’s DP‐SGD hooks expect a single, standard \textit{forward(self, input)} signature and deterministic operations. When it encounters extra arguments or stochastic operations inside the forward pass, the privacy accounting and per‐sample gradient machinery break, leading to trainer errors.

To restore compatibility, we override the model’s \textit{forward} by accepting a single concatenated tensor \textit{concat\_tensor}. Before computing the forward pass, we externally concatenate the expanded labels and the inputs along the feature dimension. Inside the new \textit{forward}, we unpack \textit{label\_B} and \textit{x\_BLCv\_wo\_first\_l} by slicing \textit{concat\_tensor}, thereby presenting Opacus with the canonical single argument signature it requires.

Once unpacked, we replicate the original embedding, positional‐encoding, and attention‐bias computations exactly as before. The unpacked \textit{label\_B} drives the class embeddings and start‐of‐sequence token, while \textit{x\_BLCv\_wo\_first\_l} populates the remainder of the input sequence. We then proceed through the usual per‐layer AdaLN self‐attention and final logits projection unmodified, ensuring functional parity with the upstream VAR implementation.